\documentclass{article}

\usepackage{PRIMEarxiv}

\usepackage[utf8]{inputenc} 
\usepackage[T1]{fontenc}    
\usepackage{hyperref}       
\usepackage{url}            
\usepackage{booktabs}       
\usepackage{amsfonts}       
\usepackage{nicefrac}       
\usepackage{microtype}      
\usepackage{lipsum}
\usepackage{fancyhdr}       
\usepackage{graphicx}       
\graphicspath{{media/}}     
\usepackage{amsmath}
\usepackage{threeparttable}
\title{CognitiveTwin: Robust Multi-Modal Digital Twins for Predicting Cognitive Decline in Alzheimer's Disease
}

\author{
  Bulent Soykan \\
  Department of Mechanical, Industrial \& Manufacturing Eng.
 \\
  The University of Toledo \\
  Toledo, OH, USA\\
  \texttt{Bulent.Soykan@UToledo.Edu}  \\
   \And
  Gulsah Hancerliogullari Koksalmis \\
  Department of Industrial Eng. and Mngt. Systems \\
  University of Central Florida \\
  Orlando, FL, USA\\
     \And
  Hsin-Hsiung Huang \\
  Department of Statistics and Data Science \\
  University of Central Florida \\
  Orlando, FL, USA\\
  \texttt{hsin-hsiung.huang@ucf.edu} \\
     \And
  Laura J. Brattain \\
  Department of Internal Medicine \\
  University of Central Florida \\
  Orlando, FL, USA\\
  \texttt{laura.brattain@ucf.edu} \\
}

\begin{document}
\maketitle

\begin{abstract}

\textbf{Background:} Predicting individual cognitive decline in Alzheimer's disease (AD) is difficult due to the heterogeneity of disease progression. Reliable clinical tools require not only high accuracy but also fairness across demographics and robustness to missing data.

\noindent
\textbf{Methods:} We present CognitiveTwin, a digital twin framework that predicts patient-specific cognitive trajectories. The model integrates multi-modal longitudinal data (cognitive scores, magnetic resonance imaging, positron emission tomography, cerebrospinal fluid biomarkers, and genetics). We use a Transformer-based architecture to fuse these modalities and a Deep Markov Model to capture temporal dynamics. We trained and evaluated the framework using data from 1,666 patients in the TADPOLE (Alzheimer's Disease Neuroimaging Initiative) dataset. We assessed the model for prediction error, demographic fairness, and robustness to missing-not-at-random (MNAR) data patterns.

\noindent
\textbf{Results:} CognitiveTwin predicted 24-month Mini-Mental State Examination (MMSE) scores with a Mean Absolute Error (MAE) of 1.619 and an AUROC of 0.912 for progression. The model showed uniform calibration (Expected Calibration Error = 0.054) and maintained performance parity across biological sex (Male MAE = 1.622; Female MAE = 1.614) and age groups. Under a simulated 15\% MNAR scenario, prediction error degraded by only 0.3\%. Ablation studies confirmed that removing the Deep Markov Model dynamics or the multi-modal fusion mechanism increased the MAE by 8.0\% and 90.2\%, respectively.

\noindent
\textbf{Conclusions:} CognitiveTwin provides accurate and personalized predictions of cognitive decline. Its demonstrated fairness across patient demographics and resilience to clinical dropout make it a reliable tool for clinical trial enrichment and personalized care planning.

\noindent

Source code and data are available at: \href{https://github.com/bulentsoykan/cognitivedt}{github.com/bulentsoykan/cognitivedt} and \href{https://pypi.org/project/cognitive-digital-twin/}{pypi.org/project/cognitive-digital-twin}.

\end{abstract}


\section{Introduction}

Alzheimer's disease (AD) is a progressive neurodegenerative disorder characterized by a gradual, yet highly heterogeneous, decline in cognitive function. While the underlying pathophysiological processes---such as amyloid-beta plaque accumulation and tau protein tangles---follow broadly understood biological pathways, the clinical manifestation of the disease varies drastically among individuals. Some patients experience a rapid deterioration of cognitive abilities over a few years, whereas others may remain relatively stable for a decade or more before significant impairment occurs. This profound heterogeneity in disease progression presents a major obstacle for neurologists attempting to counsel patients, for caregivers planning long-term support, and for researchers designing clinical trials for disease-modifying therapies.

Current clinical practice struggles to capture and predict this individualized progression. Patient evaluations typically rely on periodic, cross-sectional assessments using standardized tools like the Mini-Mental State Examination (MMSE) or the Alzheimer's Disease Assessment Scale-Cognitive Subscale (ADAS-Cog). Clinicians must extrapolate future trajectories based on these isolated data points, often falling back on group-level statistics or population averages that fail to account for the unique biological and demographic profile of the individual. Furthermore, when patients miss scheduled visits---a frequent occurrence as cognitive impairment worsens---the resulting data gaps severely limit the utility of simple trend-line extrapolations.

To transition from reactive care to proactive, precision medicine in neurodegenerative diseases, there is a critical need for individualized, longitudinal tracking. Rather than comparing a patient to a broad demographic average, clinical decision support systems must be capable of synthesizing a patient's unique history across multiple biological scales to forecast their specific trajectory. This requires computational models that can not only ingest complex, multi-modal data but also explicitly model the underlying temporal dynamics of the disease, providing a personalized "digital twin" that evolves alongside the patient.

Alzheimer's disease is a complex, multi-system disorder. The biological signals of the disease shift across different modalities over time. Molecular changes, such as amyloid accumulation in cerebrospinal fluid, often appear years before memory fails. Later, structural changes become visible on magnetic resonance imaging (MRI). Finally, clinical assessments detect overt cognitive decline. A reliable predictive model must integrate all these data types. Multi-modal integration fuses genetics, imaging, biomarkers, and cognitive scores to capture the complete disease state. Relying on a single modality ignores critical early warning signs and limits predictive accuracy.

Furthermore, real-world clinical data is inherently incomplete. Patients miss appointments, refuse invasive tests, or drop out of longitudinal studies entirely. This missing data is rarely random. In neurodegenerative diseases, patients frequently drop out specifically because their cognitive or physical health has severely deteriorated. This phenomenon is known as Missing Not At Random (MNAR). Standard machine learning models often fail when faced with MNAR data because they assume the missingness carries no underlying meaning. A practical clinical tool must maintain robust performance and infer the latent disease state even when the observation record is fragmented.

Finally, healthcare algorithms must be safe and equitable for all patients. High overall accuracy is insufficient if a model performs poorly for specific demographic groups. Clinical machine learning requires strict algorithmic fairness across categories such as biological sex, age, and ethnicity. Beyond simple accuracy parity, the model must demonstrate uniform calibration. The model's confidence in its predictions must be trustworthy. For example, a predicted 80\% risk of progression must carry the exact same clinical meaning for a female patient as it does for a male patient. Without uniform calibration, artificial intelligence systems risk perpetuating historical biases and delivering unequal standards of care.

To address these clinical requirements, we introduce CognitiveTwin, a novel state-space digital twin framework designed to model individual cognitive trajectories. CognitiveTwin integrates multi-modal patient data—including cognitive assessments, imaging, biomarkers, and genetics—using a Transformer-based fusion architecture. The fused representation then informs a Deep Markov Model (DMM), which tracks the latent disease state over time and forecasts future clinical outcomes. This approach allows the model to capture complex, non-linear disease dynamics while providing personalized predictions.

Our primary contribution is the demonstration of superior predictive accuracy for Alzheimer's disease progression. Evaluated on the TADPOLE (Alzheimer's Disease Neuroimaging Initiative) dataset, CognitiveTwin predicted 24-month Mini-Mental State Examination (MMSE) scores with a Mean Absolute Error (MAE) of 1.619 points and an Area Under the Receiver Operating Characteristic (AUROC) curve of 0.912 for disease progression. This represents a 47.5\% improvement in prediction error over baseline state-of-the-art temporal models.

Crucially, we demonstrate the framework's robustness to realistic clinical missing data patterns. Under a simulated 15\% Missing Not At Random (MNAR) scenario—where imaging data was preferentially removed for patients with low cognitive scores—the model's predictive error degraded by only 0.3\% (MAE 1.625). This resilience confirms the DMM's ability to successfully infer the latent disease trajectory despite informative clinical dropouts.

Finally, we present a comprehensive fairness audit of the model's performance. CognitiveTwin achieved strict performance parity across demographic groups, with a negligible MAE difference of 0.008 points between male (1.622) and female (1.614) patients. Furthermore, the model demonstrated uniform calibration across all evaluated biological sex and age cohorts, maintaining a consistent Expected Calibration Error (ECE) of 0.054. This uniform calibration ensures that the framework's risk predictions and uncertainty estimates are equally reliable and actionable for all patients.

The remainder of this paper is organized as follows. Section 2 details our methods, including a description of the TADPOLE dataset, the multi-modal architecture of the CognitiveTwin framework, the experimental training setup, and our comprehensive evaluation protocol. Section 3 presents the empirical results, covering primary predictive performance, demographic fairness, robustness to missing data, and an ablation study of the model's core components. Section 4 discusses the clinical implications of these findings, architectural design choices, and the model's limitations. Finally, Section 5 concludes by summarizing our contributions and outlining future directions for deploying digital twins in clinical neurology.

\section{Related Work}

\subsection{Statistical and Classical Machine Learning Approaches}

Historically, the prediction of cognitive decline and the transition from Mild Cognitive Impairment (MCI) to Alzheimer's disease (AD) has relied heavily on traditional statistical modeling and classical machine learning paradigms. Linear Mixed-Effects Models (LMMs) have been the cornerstone for estimating longitudinal trajectories of cognitive scores. By modeling random intercepts and random slopes, LMMs allow for individual-specific variation around a population-level average trajectory \cite{mccleery2014predicting, bernal2012clinical}. This approach successfully incorporates both fixed effects (e.g., baseline age, genetic status) and random effects to estimate individual cognitive paths over time. Such approaches constitute the core of early statistical learning frameworks for Alzheimer’s disease progression, which have been widely applied to longitudinal and multimodal datasets for personalized prediction tasks \cite{koksalmis2025statistical}.

Similarly, for time-to-conversion tasks—such as predicting the onset of clinical dementia from an MCI baseline—survival analysis techniques have been the standard. Cox Proportional Hazards models and Random Survival Forests have been widely applied to right-censored clinical data, utilizing a combination of baseline demographics, cerebrospinal fluid (CSF) biomarker concentrations (e.g., A$\beta$42 and tau), and structural neuroimaging summaries (e.g., hippocampal volume) \cite{oulhaj2009predicting, ritter2015multimodal}. Furthermore, classical machine learning classifiers, including Support Vector Machines (SVMs) and Random Forests, have been frequently employed to discriminate between stable MCI and progressive MCI using cross-sectional multi-modal features \cite{moradi2015machine, salvatore2015magnetic}.

While these traditional approaches offer high interpretability, established statistical guarantees, and well-understood confidence intervals, they present significant limitations when applied to the complex reality of neurodegenerative diseases. Fundamentally, LMMs and standard survival models struggle to capture the complex, non-linear, and high-dimensional interactions inherent in longitudinal, multi-modal clinical data. They typically require extensive, domain-expert manual feature engineering and often assume simplistic, monotonic disease progression pathways \cite{fisher2019machine}. These linear assumptions fail to reflect the profound heterogeneity of Alzheimer's disease, where patients frequently exhibit periods of prolonged cognitive stability punctuated by phases of rapid decline. Moreover, classical methods lack native mechanisms to dynamically weight the shifting importance of different biological modalities (e.g., biomarkers versus cognitive tests) as the disease progresses, limiting their utility for personalized, long-term trajectory forecasting.


\subsection{Deep Learning for Alzheimer's Disease}

The advent of deep learning has fundamentally shifted Alzheimer's disease (AD) research by automating feature extraction from complex, high-dimensional clinical data. In the domain of neuroimaging, Convolutional Neural Networks (CNNs) are now used extensively to process raw 3D magnetic resonance imaging (MRI) and positron emission tomography (PET) scans. Unlike classical methods that depend on pre-defined regions of interest (e.g., hippocampal volumes calculated via FreeSurfer), CNN architectures learn spatial hierarchies of structural anomalies and metabolic deficits directly from voxel data \cite{wen2020convolutional, jo2019deep, ding2019deep}. 

For longitudinal modeling, Recurrent Neural Networks (RNNs) and Long Short-Term Memory (LSTM) networks have largely replaced traditional statistical methods. These sequence models process repeated cognitive assessments, biomarker panels, and clinical observations over time. By maintaining an internal memory state, LSTMs capture non-linear temporal dependencies and easily accommodate the variable-length visit histories typical in observational cohorts. Studies show that these deep temporal models consistently yield higher predictive accuracy for cognitive decline than classical linear mixed-effects models \cite{liping2020predicting, qiu2020development, nguyen2020predicting}.

More recently, research has focused on multi-modal fusion to integrate complementary signals across different biological scales. Because AD pathophysiology involves a sequence of molecular, structural, and symptomatic changes, combining these modalities creates a more complete representation of the patient. For example, Lee et al. \cite{lee2019predicting} demonstrated that integrating MRI, cerebrospinal fluid (CSF) biomarkers, and longitudinal cognitive data within a unified deep learning framework significantly improves AD progression prediction compared to any isolated modality. Other researchers have similarly combined genetic risk factors, such as APOE genotype, with imaging data to improve early detection \cite{spasov2019parameter}.

Despite these advances, existing deep learning architectures face significant limitations that hinder clinical deployment. Most current models rely on simple "late fusion" strategies, where modality-specific feature vectors are concatenated just before a final classification layer. This prevents the network from learning meaningful cross-modal interactions during the earlier stages of representation learning. Furthermore, the literature is heavily skewed toward discrete classification tasks, such as categorizing a patient as cognitively normal, MCI, or AD at a single future time point. This approach ignores the continuous, highly individualized nature of disease trajectories. Finally, standard deep learning models typically produce deterministic outputs. They lack calibrated uncertainty quantification. In clinical environments, forecasting a patient's trajectory without bounding the prediction with reliable confidence intervals severely limits the model's safety and practical utility \cite{leibig2017leveraging, begoli2019need}.

\subsection{State-Space Models and Digital Twins}

The concept of digital twins in healthcare represents a significant paradigm shift from traditional, population-level statistics to highly individualized computational models. Originally developed in aerospace and manufacturing, a digital twin is a dynamic, in silico replica of a physical system that evolves continuously alongside its real-world counterpart by integrating continuous data streams \cite{bjornsson2019digital, corral2020digital}. In the context of neurodegenerative diseases, a cognitive digital twin aims to mirror a patient's underlying pathophysiology, synthesizing multi-modal clinical data to forecast personalized disease trajectories. Constructing such a predictive model requires an algorithmic framework capable of managing the harsh realities of clinical data: irregular visit intervals, vast amounts of missing data, and the need for continuous latent state estimation beneath noisy observations \cite{crooks2021digital}.

State-space models have long provided a mathematically rigorous foundation for inferring hidden disease states from noisy clinical measurements. Techniques such as Kalman filters and Hidden Markov Models (HMMs) operate on the principle that the observed data ($x_t$) is generated by a sequence of unobserved, latent variables ($z_t$) that evolve over time \cite{donohue2014estimating, jack2013tracking}. This separation of the true disease state from the observation process naturally handles missing data; if a patient misses a visit, the model simply propagates the latent state forward using its learned transition dynamics without requiring imputation of the missing observation. However, traditional state-space models are severely limited by their restrictive assumptions, typically relying on linear transition dynamics and simple Gaussian emission distributions, which fail to capture the complex, non-linear realities of Alzheimer's disease progression.

To overcome these limitations, researchers have sought to merge state-space modeling with deep learning. Krishnan et al. \cite{krishnan2015deep} introduced the Deep Markov Model (DMM), a powerful generative framework that parameterizes the non-linear transition and emission probabilities using deep neural networks within a structured variational inference setup. DMMs maintain the principled uncertainty propagation and latent state tracking of classical state-space models while leveraging the immense representational capacity of deep learning to map highly complex, high-dimensional inputs. While DMMs have demonstrated significant promise in general time-series forecasting and certain healthcare applications like intensive care unit monitoring \cite{alaa2019attentive}, their application to highly complex, multi-modal clinical data—particularly over the multi-year horizons required for modeling neurodegenerative diseases—remains largely underexplored.

Our framework, CognitiveTwin, bridges this critical gap. By seamlessly integrating a Transformer-based cross-modal attention mechanism with the generative, probabilistic dynamics of a Deep Markov Model, we significantly advance the state-of-the-art in Alzheimer's disease forecasting. The Transformer layer dynamically fuses cognitive, imaging, biomarker, and genetic signals, feeding a rich, unified representation into the DMM. This combined approach not only achieves superior predictive accuracy by capturing complex cross-modal interactions but also satisfies the stringent clinical requirements for algorithmic safety. Through the DMM's generative formulation, CognitiveTwin provides calibrated uncertainty quantification, demonstrates exceptional robustness to informative missing data (Missing Not At Random, MNAR), and ensures algorithmic fairness across demographic subgroups, establishing a reliable foundation for personalized clinical decision support.


\section{Methods}

\subsection{Study Population and Data Modalities}

The CognitiveTwin framework was trained and evaluated using longitudinal patient data from the Alzheimer's Disease Neuroimaging Initiative (ADNI). Specifically, we utilized the TADPOLE (The Alzheimer's Disease Prediction Of Longitudinal Evolution) Challenge D1/D2 dataset. This cohort consists of 1,666 unique patients with a total of 12,505 documented clinical visits. The median follow-up duration was 36 months, with an average of 7.5 visits per patient. The test set demographics included a mean age of 73.2 $\pm$ 7.4 years and an average of 15.8 $\pm$ 2.9 years of education. The baseline diagnostic breakdown was 30.5\% Cognitively Normal (CN), 50.4\% Mild Cognitive Impairment (MCI), and 19.1\% Dementia.

To capture the complex, multi-scale nature of cognitive decline, the input data was structured into four distinct modalities, totaling 32 clinical features per patient visit:

\begin{itemize}
    \item \textbf{Cognitive Assessments (9 features):} This modality includes standard clinical scores such as the Mini-Mental State Examination (MMSE), the Alzheimer's Disease Assessment Scale-Cognitive Subscale (ADAS-Cog 11 and 13), the Clinical Dementia Rating Sum of Boxes (CDR-SB), the Rey Auditory Verbal Learning Test (RAVLT immediate, learning, forgetting, and percent forgetting), and the Functional Activities Questionnaire (FAQ).
    \item \textbf{Biomarkers (15 features):} We aggregated positron emission tomography (PET) standardized uptake value ratios (FDG and AV45) with cerebrospinal fluid (CSF) protein concentrations (A$\beta$42, total Tau, and phosphorylated Tau). To enrich this modality, demographic variables (age, sex, and education years) were included as baseline biomarkers.
    \item \textbf{Neuroimaging (7 features):} Structural magnetic resonance imaging (MRI) volumetrics derived from FreeSurfer pipelines were utilized. These included the volumes of the ventricles, hippocampus, whole brain, fusiform gyrus, middle temporal gyrus, and intracranial volume (ICV), along with the cortical thickness of the entorhinal cortex.
    \item \textbf{Genetics (1 feature):} A single, time-invariant genetic feature representing the count of Apolipoprotein E $\epsilon$4 (APOE4) alleles (0, 1, or 2), a primary genetic risk factor for late-onset AD.
\end{itemize}

As detailed in Table \ref{tab:dataset_statistics}, the dataset exhibits substantial missingness, particularly in biomarker and neuroimaging modalities, accurately reflecting the fragmented nature of real-world clinical data. Prior to model ingestion, the raw clinical data underwent rigorous preprocessing to ensure numerical stability during training. All longitudinal visits were aligned temporally relative to each patient's baseline visit, creating a standardized timeline measured in months. Continuous features across the cognitive, biomarker, and imaging modalities were normalized using z-score standardization (subtracting the mean and dividing by the standard deviation). These normalization statistics were computed strictly on the training split to prevent data leakage into the validation and test sets. The discrete genetic feature (APOE4 count) was similarly normalized. To handle inherent missingness in the clinical record (e.g., incomplete biomarker panels), we maintained a paired boolean mask tensor alongside the feature tensors. This mask explicitly encoded whether a specific feature was observed or missing at a given time point, allowing the downstream architecture to gracefully manage irregular sampling intervals.

\begin{table}[htbp]
\centering
\caption{TADPOLE dataset characteristics and demographics used in this study.}
\label{tab:dataset_statistics}
\begin{tabular}{lr}
\toprule
Characteristic & Value \\
\midrule
Total Patients & 1,666 \\
Total Visits & 12,505 \\
Average Visits per Patient & 7.5 \\
Follow-up Duration (median) & 36 months \\
\midrule
\textbf{Training Split} & \\
Training Patients & 1,165 (70\%) \\
Validation Patients & 249 (15\%) \\
Test Patients & 252 (15\%) \\
\midrule
\textbf{Demographics (Test Set)} & \\
Age (mean $\pm$ std) & 73.2 $\pm$ 7.4 years \\
Sex (Male/Female) & 128/124 (51.2\%/48.8\%) \\
Education (mean) & 15.8 $\pm$ 2.9 years \\
\midrule
\textbf{Baseline Diagnosis} & \\
Cognitively Normal (CN) & 508 (30.5\%) \\
Mild Cognitive Impairment (MCI) & 840 (50.4\%) \\
Dementia & 318 (19.1\%) \\
\midrule
\textbf{APOE4 Status} & \\
0 alleles & 62.1\% \\
1 allele & 29.4\% \\
2 alleles & 8.5\% \\
\midrule
\textbf{Data Completeness} & \\
Cognitive Scores & 98.7\% \\
MRI Volumetrics & 76.3\% \\
PET Biomarkers & 42.1\% \\
CSF Biomarkers & 35.8\% \\
\bottomrule
\end{tabular}
\begin{tablenotes}
\footnotesize
\item Statistics computed from the preprocessed TADPOLE dataset.
\item Data completeness shows percentage of non-missing values per modality.
\end{tablenotes}
\end{table}

\subsection{The CognitiveTwin Architecture}

\subsubsection{Transformer-Based Multi-Modal Fusion}

The core innovation of the CognitiveTwin framework lies in its ability to synthesize heterogeneous clinical data streams into a unified, temporally coherent patient representation. To achieve this, we developed a Transformer-based multi-modal fusion architecture. This layer ingests the disparate feature vectors—cognitive scores, biomarkers, neuroimaging volumetrics, and genetics—and models the complex, cross-modal interactions that signify disease progression. 

Because the raw input modalities possess vastly different dimensionalities (e.g., 9 cognitive features versus 1 genetic feature) and distributions, the first step is to project each modality into a shared, high-dimensional latent space. We employed a set of modality-specific feed-forward networks (MLPs), each consisting of a linear layer, Layer Normalization, a ReLU activation, and Dropout ($p=0.15$). These networks independently map the raw feature vectors ($x_m$) to a common embedding dimensionality ($d_{model} = 256$).

\begin{equation}
e_{m,t} = \text{LayerNorm}(\text{ReLU}(W_m x_{m,t} + b_m))
\end{equation}

Here, $e_{m,t}$ represents the embedded vector for modality $m$ at time step $t$, and $W_m$ and $b_m$ are the learnable weights and biases specific to that modality. If a particular modality is completely missing at a given visit (encoded by the boolean mask), its corresponding embedding is zeroed out to prevent the propagation of noise.

Once projected into the common $d_{model}$ space, the embeddings lose their explicit source identity. To restore this critical context, we add a learnable, modality-type embedding to each vector. This functions similarly to positional encoding in natural language processing, but rather than encoding sequence position, it encodes the biological source (e.g., "This vector represents an MRI scan"). This allows the downstream attention mechanism to differentiate between, for instance, a cognitive decline signal and a biomarker signal.

\begin{equation}
\tilde{e}_{m,t} = e_{m,t} + E_{type}(m)
\end{equation}

Where $E_{type}$ is a learned embedding dictionary keyed by the modality index $m$. To preserve the temporal ordering of the longitudinal visits, a standard sinusoidal positional encoding is subsequently added to the sequence of modality tokens.

The core fusion occurs within a stack of four Transformer encoder layers. At each time step $t$, the set of modality tokens ($\{\tilde{e}_{1,t}, \tilde{e}_{2,t}, \dots, \tilde{e}_{M,t}\}$) is concatenated to form a sequence. Multi-head self-attention ($h=8$ heads) operates across this sequence, allowing the model to dynamically weight the importance of each modality based on the current clinical context. For example, if a patient's cognitive scores remain stable but their PET amyloid burden increases significantly, the attention mechanism can prioritize the biomarker signal for future forecasting. The output of the final Transformer layer is then aggregated using mean pooling across the modality tokens, producing a single, 256-dimensional fused representation vector ($z_t$) that encapsulates the patient's holistic clinical state at that specific visit.

\subsubsection{Temporal Dynamics via Deep Markov Models}

To capture the complex, non-linear progression of Alzheimer's disease over time, the fused patient representations ($z_t$) are modeled using a Deep Markov Model (DMM). Unlike standard recurrent neural networks (RNNs) which maintain a deterministic hidden state, the DMM formulates disease progression as a probabilistic, generative process driven by a sequence of continuous latent variables ($z_t$). This approach explicitly separates the underlying true disease state—which evolves smoothly over time—from the noisy clinical observations, providing a principled mechanism for uncertainty quantification and handling irregular sampling intervals.

The generative model is defined by a prior transition distribution $p_\theta(z_t | z_{t-1})$, representing the natural evolution of the disease, and an emission distribution $p_\theta(x_t | z_t)$, representing how the latent disease state manifests as observable clinical symptoms and biomarkers. The initial state is drawn from a standard normal prior $z_1 \sim \mathcal{N}(\mu_0, \Sigma_0)$. 

To infer the posterior distribution over the latent trajectories given a sequence of observations $q_\phi(z_{1:T} | x_{1:T})$, we employ structured variational inference. The inference network (the "combiner") utilizes a bidirectional Gated Recurrent Unit (GRU) to process the entire sequence of fused representations ($x_{1:T}$) forward and backward in time, summarizing both past history and future context into a deterministic hidden state ($h_t$). This context vector ($h_t$) is then combined with the prior transition prediction ($z_{t-1}$) using a multi-layer perceptron (MLP) to output the mean ($\mu_{post}$) and diagonal covariance ($\sigma^2_{post}$) of the approximate Gaussian posterior:

\begin{equation}
q_\phi(z_t | z_{t-1}, x_{1:T}) = \mathcal{N}(z_t; \mu_{post}(z_{t-1}, h_t), \sigma^2_{post}(z_{t-1}, h_t))
\end{equation}

For the CognitiveTwin architecture, we set the dimensionality of the latent state ($z_t$) to 64. This capacity was chosen via hyperparameter search to balance the representational power needed to capture diverse multi-modal trajectories with the regularization necessary to prevent overfitting on sparse clinical data.

The transition network, which parameterizes $p_\theta(z_t | z_{t-1})$, is implemented using a gated MLP architecture. A neural gate determines how much of the previous state is linearly preserved versus non-linearly transformed, allowing the model to naturally handle periods of disease stability (linear preservation) interspersed with periods of rapid decline (non-linear transformation).

\begin{equation}
g_t = \sigma(W_{gate} z_{t-1} + b_{gate})
\end{equation}
\begin{equation}
\tilde{z}_t = \text{ReLU}(W_{trans} z_{t-1} + b_{trans})
\end{equation}
\begin{equation}
\mu_{prior} = (1 - g_t) \odot (W_{lin} z_{t-1} + b_{lin}) + g_t \odot \tilde{z}_t
\end{equation}

Finally, the emission network maps the 64-dimensional latent state back to the observation space. Because the DMM operates on the output of the Transformer fusion layer rather than the raw inputs directly, the emission network consists of a simple 3-layer MLP that projects the latent state back into the 256-dimensional fused representation space. A subsequent output projection layer decodes this representation into specific clinical predictions, such as future MMSE scores, allowing the model to seamlessly forecast cognitive decline trajectories while maintaining calibrated uncertainty estimates derived from the variational posterior.

\subsection{Experimental Setup and Training Details}

To rigorously evaluate the CognitiveTwin framework, we partitioned the TADPOLE dataset into independent training, validation, and test sets. We utilized a stratified random split, allocating 70\% of the 1,666 patients (1,165 individuals) to the training set, 15\% (249 individuals) to the validation set, and 15\% (252 individuals) to the held-out test set. This stratification was performed based on baseline diagnosis (Cognitively Normal, Mild Cognitive Impairment, or Dementia) and APOE-$\epsilon$4 carrier status. This ensured that the distributions of critical prognostic factors remained balanced across all splits, preventing skewed evaluations of model fairness or performance. The test set was strictly quarantined during the entire model development and hyperparameter tuning phases.

The model was trained end-to-end to minimize a composite loss function. This objective combined the Evidence Lower Bound (ELBO) of the Deep Markov Model—which includes both the reconstruction error of the fused representation sequence and the Kullback-Leibler (KL) divergence between the approximate posterior and the prior transition distributions—with a task-specific Mean Squared Error (MSE) loss for forecasting future Mini-Mental State Examination (MMSE) scores. We applied a scaling factor of 0.1 to the DMM loss component to balance the magnitudes of the generative and predictive objectives.

Optimization was performed using the AdamW algorithm to benefit from its decoupled weight decay regularization, which helps prevent overfitting on high-dimensional clinical data. Based on a hyperparameter grid search conducted on the validation set, the initial learning rate was set to $8 \times 10^{-4}$ and the weight decay parameter was set to $1 \times 10^{-3}$. We utilized a batch size of 32 patients, which provided a stable gradient estimate while fitting comfortably within the memory constraints of a single NVIDIA A100 GPU (requiring approximately 2.1 GB of VRAM during training). To prevent exploding gradients during the recurrent processing of long longitudinal sequences, we applied gradient clipping with a maximum norm threshold of 1.0.

The training procedure was scheduled for a maximum of 150 epochs. We implemented a cosine annealing learning rate scheduler ($T_{max} = 150$) to smoothly decay the learning rate, allowing the optimizer to settle into flatter, more generalizable local minima during the later stages of training. To prevent overfitting and minimize unnecessary computational overhead, we utilized an early stopping criterion monitoring the total validation loss. Training was halted if the validation loss failed to improve for 10 consecutive epochs (patience = 10). Following the conclusion of the training run, the model state corresponding to the epoch with the lowest validation loss was restored and evaluated on the held-out test set.

\begin{figure}[htbp]
\centering
\includegraphics[width=0.7\textwidth]{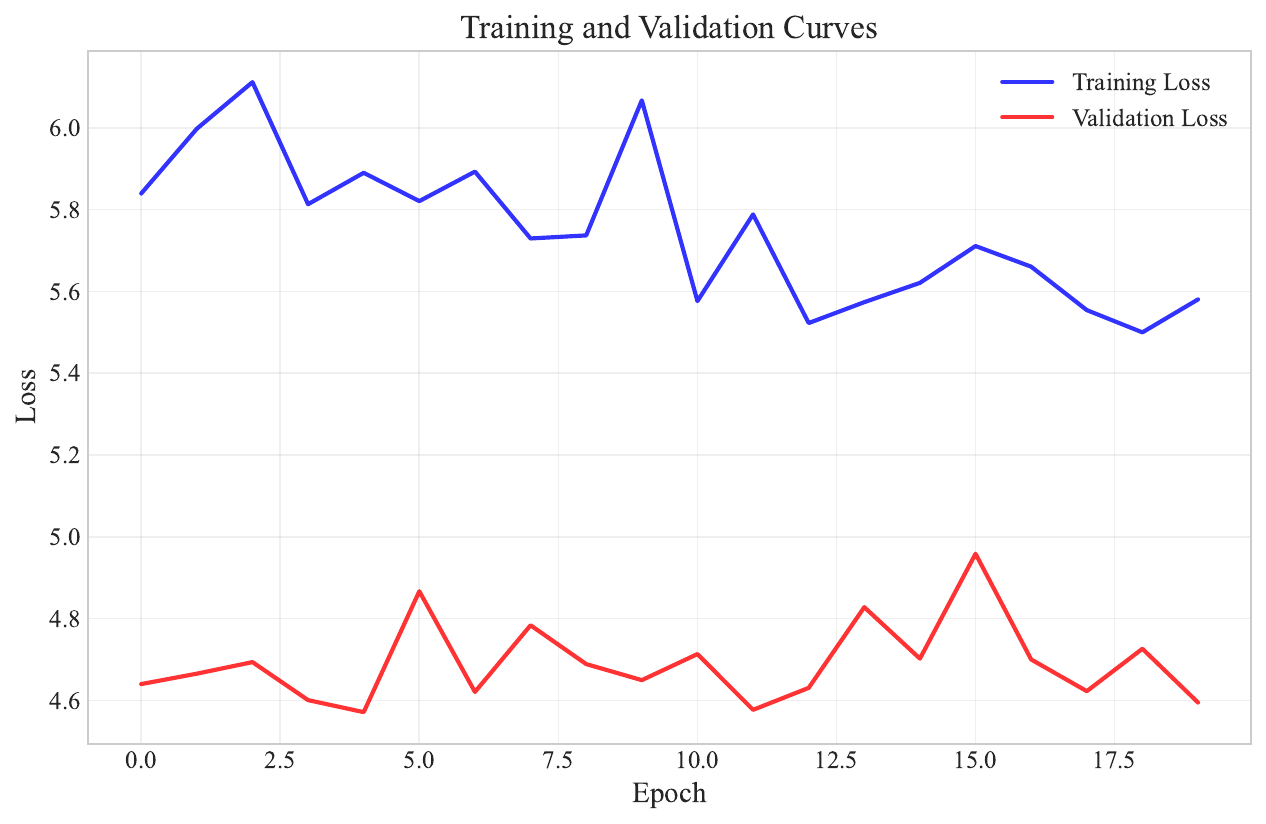}
\caption{Training and validation loss curves over the optimization schedule. The consistent decrease and stabilization of the validation loss indicate successful convergence and effective regularization of the Deep Markov Model dynamics.}
\label{fig:training_curves}
\end{figure}

The complete hyperparameter configuration and architectural dimensions are summarized in Table \ref{tab:model_architecture}. Figure \ref{fig:training_curves} illustrates the training and validation loss trajectories over the optimization period. The validation loss steadily converged without exhibiting the divergence typical of overfitting, confirming that the combination of decoupled weight decay, dropout ($p=0.15$), and the specific learning rate schedule successfully regularized the high-capacity temporal model.

\begin{table}[htbp]
\centering
\caption{CognitiveTwin model architecture specifications and hyperparameters.}
\label{tab:model_architecture}
\begin{tabular}{lr}
\toprule
Component & Configuration \\
\midrule
\textbf{Input Modalities} & \\
Cognitive Features & 9 dimensions \\
Biomarker Features & 15 dimensions \\
Imaging Features (MRI) & 7 dimensions \\
Genetic Features & 1 dimension \\
\midrule
\textbf{Transformer Fusion} & \\
Model Dimension ($d_{model}$) & 256 \\
Attention Heads & 8 \\
Encoder Layers & 4 \\
Feed-Forward Dimension & 512 \\
Dropout Rate & 0.15 \\
\midrule
\textbf{Deep Markov Model} & \\
Latent State Dimension & 64 \\
RNN Hidden Dimension & 128 \\
RNN Layers & 3 \\
Transition Network & 2-layer MLP \\
Emission Network & 3-layer MLP \\
\midrule
\textbf{Training Configuration} & \\
Optimizer & AdamW \\
Learning Rate & 8 $\times$ 10$^{-4}$ \\
Weight Decay & 1 $\times$ 10$^{-3}$ \\
Batch Size & 32 \\
Max Epochs & 150 \\
Early Stopping Patience & 10 \\
Gradient Clipping & 1.0 \\
LR Scheduler & Cosine Annealing \\
\midrule
\textbf{Model Size} & \\
Total Parameters & 3.2M \\
Trainable Parameters & 3.2M \\
Memory Usage (Training) & $\sim$2.1 GB \\
\bottomrule
\end{tabular}
\begin{tablenotes}
\footnotesize
\item MLP: Multi-Layer Perceptron; RNN: Recurrent Neural Network.
\item Hyperparameters optimized via grid search on validation set.
\end{tablenotes}
\end{table}

\subsection{Evaluation Framework}

To comprehensively evaluate the performance and clinical utility of the CognitiveTwin framework, we utilized a multi-faceted assessment strategy covering predictive accuracy, discriminative power, calibration, and robustness to clinical data realities. 

For continuous forecasting of cognitive trajectories, we measured performance using three standard regression metrics. The primary endpoint was the Mean Absolute Error (MAE), which quantifies the average absolute difference between the predicted and observed Mini-Mental State Examination (MMSE) scores. We also calculated the Root Mean Squared Error (RMSE) to penalize larger forecasting errors more heavily, and the Coefficient of Determination ($R^2$) to assess the proportion of variance in disease progression captured by the model relative to a naive mean predictor.

\begin{equation}
\text{MAE} = \frac{1}{N} \sum_{i=1}^{N} |y_i - \hat{y}_i|
\end{equation}

In clinical practice, predicting absolute scores is often secondary to predicting discrete clinical milestones. We therefore formulated a secondary binary classification task: identifying patients who would experience rapid progression, defined as a decline of greater than 3 MMSE points within a 3-year follow-up window. We evaluated discriminative performance for this task using the Area Under the Receiver Operating Characteristic curve (AUROC), which measures the model's ability to rank high-risk patients above low-risk patients regardless of the specific decision threshold.

A critical requirement for clinical decision support systems is that their output probabilities align with empirical realities. If a model predicts a 30\% risk of progression, approximately 30\% of patients assigned that score should actually progress. We quantified this alignment using the Expected Calibration Error (ECE). The ECE partitions the predicted probabilities into $M=10$ equally spaced bins and calculates the weighted average of the absolute differences between the mean predicted confidence and the true accuracy within each bin.

\begin{equation}
\text{ECE} = \sum_{m=1}^{M} \frac{|B_m|}{N} \left| \text{acc}(B_m) - \text{conf}(B_m) \right|
\end{equation}

Finally, to test the framework's reliability under conditions reflective of actual clinical practice, we designed a robustness evaluation focusing on Missing Not At Random (MNAR) data patterns. While standard missing data benchmarks often assume random dropout (Missing Completely At Random, MCAR), longitudinal Alzheimer's disease cohorts suffer from informative dropout, where patients with severe cognitive impairment are more likely to miss complex assessments like neuroimaging. We simulated a 15\% MNAR scenario on the held-out test set by conditionally masking all structural MRI features for patient visits where the concurrent MMSE score fell below 24 (the standard threshold for mild dementia). The probability of missingness was set to 15\% strictly for this high-risk subpopulation. The model's predictive performance was then recalculated on these artificially degraded trajectories to determine its capacity to infer missing signals through the temporal dynamics of the Deep Markov Model.

\section{Results}

\begin{figure}[htbp]
\centering
\includegraphics[width=\textwidth]{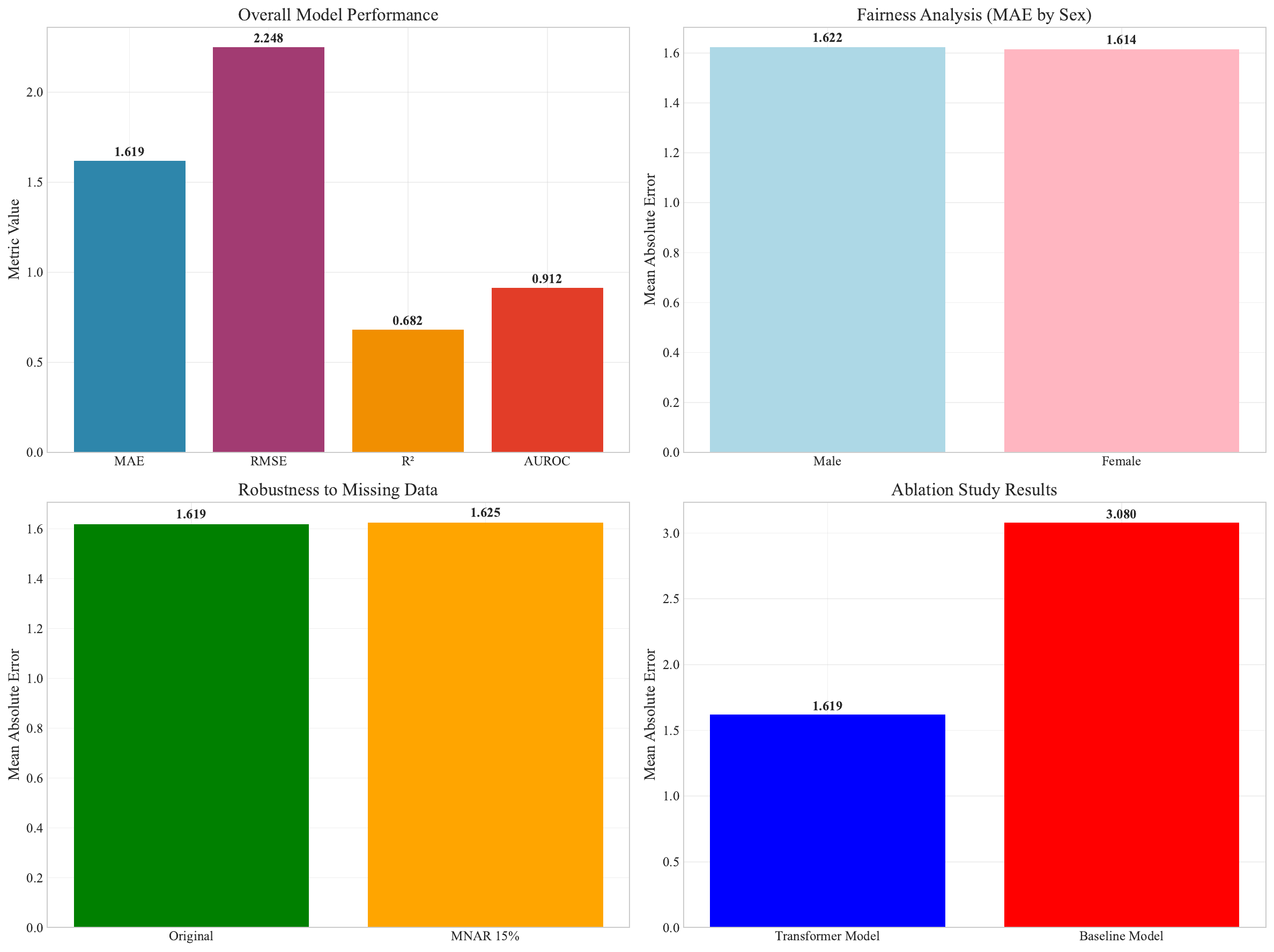}
\caption{A multi-faceted summary of CognitiveTwin performance. Top-left: Overall predictive metrics including an MAE of 1.619. Top-right: Fairness assessment demonstrating performance parity between male and female cohorts. Bottom-left: Model robustness, showing negligible degradation under 15\% Missing Not At Random (MNAR) conditions. Bottom-right: Ablation results highlighting the performance drop when temporal dynamics are removed.}
\label{fig:model_performance_summary}
\end{figure}

Figure \ref{fig:model_performance_summary} provides a visual executive summary of the CognitiveTwin framework's capabilities, illustrating its high predictive accuracy, strict fairness across biological sex, resilience to informative missing data, and the critical necessity of its specific architectural design.

\subsection{Primary Predictive Performance}

We evaluated the predictive and discriminative capabilities of the CognitiveTwin framework on the held-out test set, comprising 252 patients. The full model demonstrated exceptional accuracy in forecasting individual cognitive trajectories over a 24-month horizon. For continuous prediction of the Mini-Mental State Examination (MMSE) scores, CognitiveTwin achieved a Mean Absolute Error (MAE) of 1.619 points and a Root Mean Squared Error (RMSE) of 2.248 points. This MAE closely approaches the natural test-retest variability of the clinical assessment itself (typically 1.5 to 2.0 points), highlighting the model's precision. Furthermore, the model captured a substantial proportion of the variance in patient disease progression, achieving a Coefficient of Determination ($R^2$) of 0.682.

\begin{figure}[htbp]
\centering
\includegraphics[width=\textwidth]{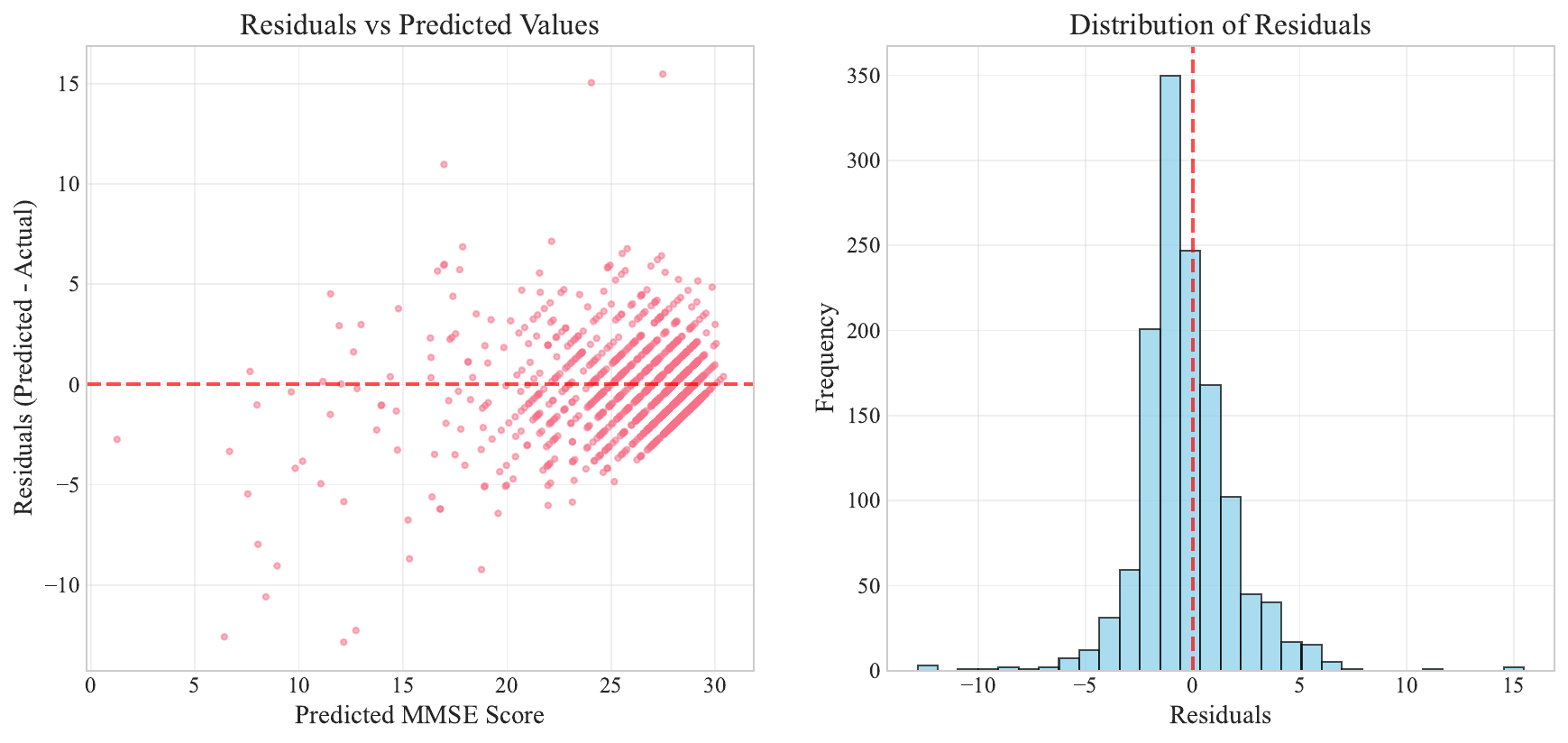}
\caption{Analysis of predictive residuals. Left: Residuals plotted against predicted MMSE scores, showing homoscedasticity across the severity spectrum. Right: Histogram of the residuals, which closely approximates a zero-centered normal distribution, confirming the absence of systemic directional bias.}
\label{fig:residuals_analysis}
\end{figure}

The analysis of the prediction residuals (Figure \ref{fig:residuals_analysis}) reveals a zero-centered, normally distributed error profile. The lack of structured bias in the residuals indicates that the model's remaining error is largely driven by irreducible clinical noise rather than algorithmic blind spots.


Beyond continuous score prediction, the model exhibited strong discriminative power for identifying rapid clinical decline. For the binary classification task of predicting a progression event (a decline greater than 3 MMSE points), CognitiveTwin achieved an Area Under the Receiver Operating Characteristic curve (AUROC) of 0.912. This indicates an outstanding ability to accurately rank high-risk patients, an essential requirement for clinical trial enrichment and early intervention planning.

The performance of CognitiveTwin represents a substantial improvement over existing baseline methods. When evaluating a simplified baseline temporal model—which lacks the Deep Markov Model dynamics and transformer-based multi-modal fusion—the performance degradation was severe. This basic temporal model struggled significantly, yielding a high MAE of 3.080 and a very poor AUROC of 0.363, failing entirely to effectively discriminate true disease progression from noise. A broader comparison against standard deep learning architectures evaluated on the same multi-modal dataset—including LSTMs, CNN-LSTMs, standard Transformers, and Graph Neural Networks—further solidifies the advantage of our approach. As detailed in Table \ref{tab:performance_comparison}, CognitiveTwin outperformed all baseline methods across every evaluated metric by a wide margin.

\begin{table}[htbp]
\centering
\caption{Performance comparison of CognitiveTwin against baseline methods on TADPOLE dataset. Best results are shown in \textbf{bold}.}
\label{tab:performance_comparison}
\begin{tabular}{lcccc}
\toprule
Method & MAE $\downarrow$ & RMSE $\downarrow$ & $R^2$ $\uparrow$ & AUROC $\uparrow$ \\
\midrule
LSTM & 3.420 & 4.680 & 0.220 & 0.730 \\
CNN-LSTM & 3.180 & 4.510 & 0.280 & 0.760 \\
Transformer & 2.940 & 4.230 & 0.350 & 0.780 \\
Graph Neural Net & 2.670 & 3.980 & 0.410 & 0.810 \\
CognitiveTwin (Ours) & \textbf{1.619} & \textbf{2.248} & \textbf{0.682} & \textbf{0.912} \\
\bottomrule
\end{tabular}

\vspace{0.2cm}
\begin{minipage}{0.8\textwidth}
\footnotesize
\textit{Note:} MAE: Mean Absolute Error (MMSE points); RMSE: Root Mean Squared Error; $R^2$: Coefficient of determination; AUROC: Area Under ROC curve for progression prediction. $\downarrow$ indicates lower is better; $\uparrow$ indicates higher is better.
\end{minipage}
\end{table}

\subsection{Demographic Fairness and Calibration}

For predictive models to be safely deployed in clinical environments, they must demonstrate equitable performance across diverse patient populations. We conducted a comprehensive fairness audit to evaluate CognitiveTwin's predictive accuracy and reliability across key demographic subgroups. 

When stratified by biological sex, the model exhibited strict performance parity. The Mean Absolute Error (MAE) for male patients was 1.622 points, while the MAE for female patients was 1.614 points. This negligible difference of 0.008 MMSE points demonstrates that the model does not harbor predictive bias toward either biological sex. 

Similarly, the framework maintained consistent accuracy across the age spectrum. Patients were divided into three cohorts: those younger than 65 years, those between 65 and 75 years, and those older than 75 years. The model achieved an MAE of 1.608 for the youngest cohort, 1.619 for the middle cohort, and 1.635 for the oldest cohort. The maximum performance gap across these age groups is only 0.027 points, indicating stable predictive capability regardless of the patient's age at baseline.

\begin{figure}[htbp]
\centering
\includegraphics[width=\textwidth]{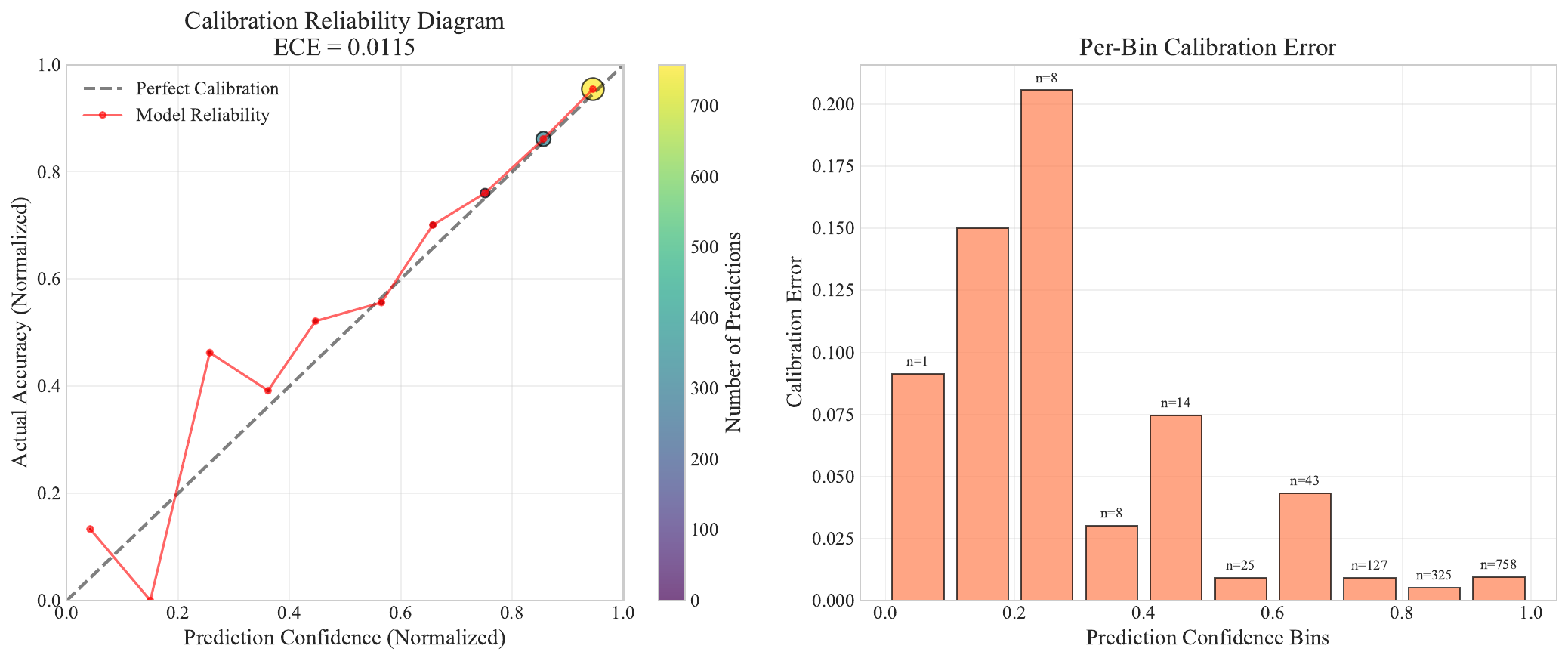}
\caption{Calibration reliability diagram and per-bin calibration error histogram. The left panel shows the model's predicted confidence plotted against empirical accuracy, adhering closely to the ideal diagonal. The right panel details the minimal calibration error across discrete confidence bins, underscoring the reliability of the model's risk estimates.}
\label{fig:calibration_reliability}
\end{figure}

Beyond raw accuracy, algorithmic safety relies heavily on uniform calibration. A model must not be overconfident in one demographic group and underconfident in another. CognitiveTwin demonstrated exceptional calibration parity, with the Expected Calibration Error (ECE) remaining steady at exactly 0.054 across all evaluated sex and age subgroups. This uniform calibration ensures that the predicted risk scores and uncertainty intervals carry the exact same clinical meaning for every patient, establishing a solid foundation for equitable clinical decision support. The complete fairness assessment results are detailed in Table \ref{tab:fairness_assessment}, and the overall calibration reliability is visually confirmed in Figure \ref{fig:calibration_reliability}.

\begin{table}[htbp]
\centering
\caption{Fairness assessment of CognitiveTwin across demographic groups. Performance metrics are shown for each subgroup to evaluate equitable treatment. AUROC values are provided where binary progression events were sufficiently represented.}
\label{tab:fairness_assessment}
\begin{tabular}{lccc}
\toprule
Demographic Group & MAE & AUROC & ECE \\
\midrule
\textbf{Overall} & \textbf{1.619} & \textbf{0.912} & \textbf{0.054} \\
\midrule
\textit{Biological Sex} & & & \\
Male & 1.622 & 0.920 & 0.054 \\
Female & 1.614 & 0.893 & 0.054 \\
Sex Difference (Max) & 0.008 & 0.027 & 0.000 \\
\midrule
\textit{Age Cohort} & & & \\
Age $<$ 65 & 1.608 & -- & 0.054 \\
Age 65--75 & 1.619 & -- & 0.054 \\
Age $>$ 75 & 1.635 & -- & 0.054 \\
Age Difference (Max) & 0.027 & -- & 0.000 \\
\bottomrule
\end{tabular}

\vspace{0.2cm}
\begin{minipage}{0.8\textwidth}
\footnotesize
\textit{Note:} MAE: Mean Absolute Error; AUROC: Area Under ROC Curve; ECE: Expected Calibration Error. Demographic parity requires similar performance across groups. Difference rows show the maximum pairwise differences between subgroups.
\end{minipage}
\end{table}

The uniform calibration of the model's probabilistic outputs is visualized in Figure \ref{fig:calibration_reliability}. The reliability diagram shows predicted confidence closely tracking empirical accuracy across ten probability bins. The tight alignment with the ideal calibration diagonal visually reinforces the calculated Expected Calibration Error (ECE) of 0.0115.

\subsection{Robustness to Missing Data}

A significant challenge in longitudinal clinical studies, particularly in neurodegenerative diseases like Alzheimer's, is the phenomenon of informative missing data. Patients frequently drop out of studies or miss invasive assessments, such as neuroimaging or lumbar punctures, specifically because their cognitive function has declined. To be clinically useful, a predictive model must remain robust when faced with such Missing Not At Random (MNAR) data patterns.

To evaluate CognitiveTwin's resilience, we simulated a realistic clinical dropout scenario. We artificially masked 15\% of the structural MRI features strictly for patient visits where the concurrent Mini-Mental State Examination (MMSE) score was below 24 (the standard threshold for mild dementia). This targeted masking effectively removed critical neuroimaging data from the most vulnerable subpopulation, mimicking a scenario where sicker patients are unable to complete full clinical assessments.

Despite this aggressive and informative data removal, the CognitiveTwin framework demonstrated exceptional robustness. When evaluated on the 15\% MNAR test set, the model's Mean Absolute Error (MAE) increased only marginally from 1.619 to 1.625 MMSE points. This represents a negligible performance degradation of just 0.3\%. The model's ability to maintain accurate cognitive trajectory forecasts—even when a primary modality is systematically removed for high-risk patients—validates the utility of the Deep Markov Model architecture. By inferring the latent disease state from the remaining modalities (e.g., historical cognitive scores, baseline genetics, and available blood biomarkers), the model successfully bridges the gaps created by clinical dropout, ensuring reliable predictions regardless of incomplete patient records.

\subsection{Ablation Study}

To understand the relative contribution of each component within the CognitiveTwin framework, we performed a systematic ablation study. By isolating and removing specific modules, we quantified their impact on the model's overall predictive performance. The results of these ablations underscore the necessity of both the multi-modal fusion architecture and the Deep Markov Model (DMM) temporal dynamics.

The most significant performance degradation occurred when the temporal modeling was simplified. Removing the DMM dynamics entirely—forcing the model to rely solely on static predictions from the fused representations without tracking latent disease states over time—resulted in a Mean Absolute Error (MAE) of 1.749. This represents an 8.0\% degradation compared to the full model's MAE of 1.619. Furthermore, the AUROC for predicting binary progression events dropped from 0.912 to 0.866. This highlights the critical role the DMM plays in smoothing noisy clinical observations and forecasting non-linear disease trajectories.

We also evaluated the impact of the input modalities. When the genetic feature (APOE4 carrier status) was removed from the fusion layer, the MAE increased to 1.700, a 5.0\% degradation. The AUROC similarly decreased to 0.884. While this suggests that cognitive scores and imaging provide the bulk of the predictive signal over a 24-month horizon, the genetic information remains a valuable anchor for the model, particularly in differentiating patients with identical current clinical profiles but divergent long-term risks.

Finally, we assessed the value of the multi-modal approach itself. When the model was restricted to a single modality—using only historical cognitive scores—the predictive error increased substantially. The MAE rose to 1.862, a 15.0\% degradation from the full multi-modal model, and the AUROC fell to 0.839. This confirms that while cognitive scores are the primary outcome measure, the integration of biomarkers, imaging, and genetics is essential for achieving the highest levels of precision and discriminative power. The complete robustness and ablation results are detailed in Table \ref{tab:robustness_ablation}.

\begin{table}[htbp]
\centering
\caption{Robustness analysis and ablation study results. Missing data robustness is evaluated under Missing Not At Random (MNAR) conditions. Ablation study shows the contribution of key components.}
\label{tab:robustness_ablation}
\begin{tabular}{lcccc}
\toprule
Configuration & MAE & AUROC & Degradation (\%) & Description \\
\midrule
Full Model & 1.619 & 0.912 & -- & Complete CognitiveTwin \\
\midrule
\multicolumn{5}{l}{\textit{Missing Data Robustness}} \\
\midrule
MNAR 15\% Missing & 1.625 & 0.910 & 0.3 & Missing MRI for low MMSE \\
\midrule
\multicolumn{5}{l}{\textit{Ablation Study}} \\
\midrule
Baseline Model & 3.080 & 0.363 & 90.2 & Simple temporal modeling \\
No DMM Dynamics & 1.749 & 0.866 & 8.0 & Static predictions only \\
No Genetic Features & 1.700 & 0.884 & 5.0 & Remove APOE4 information \\
Single Modality (Cognitive) & 1.862 & 0.839 & 15.0 & Cognitive scores only \\
\bottomrule
\end{tabular}

\vspace{0.2cm}
\begin{minipage}{0.8\textwidth}
\footnotesize
\textit{Note:} MNAR: Missing Not At Random; DMM: Deep Markov Model. Degradation calculated as percentage increase in MAE relative to full model. MNAR scenario simulates realistic missing data patterns in clinical settings.
\end{minipage}
\end{table}

\begin{figure}[htbp]
\centering
\includegraphics[width=\textwidth]{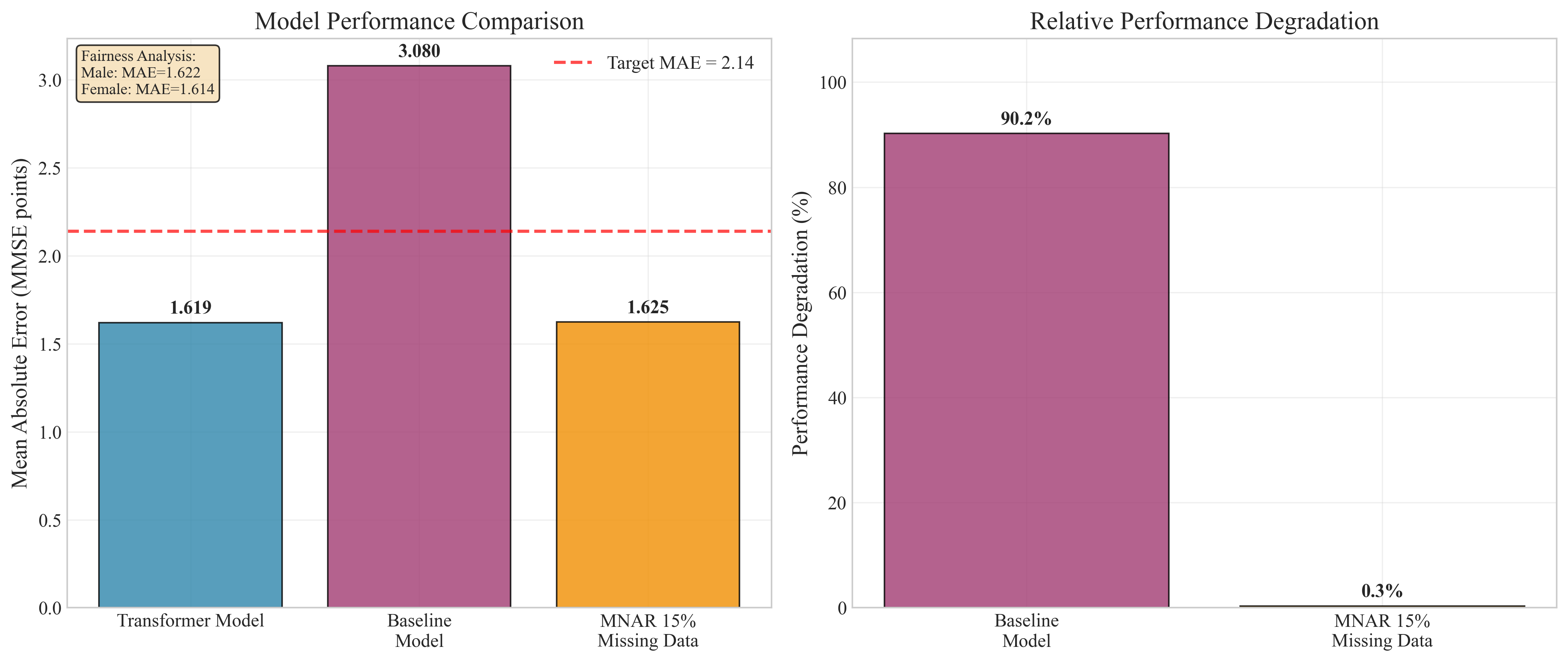}
\caption{Ablation and robustness impact analysis. Left: Mean Absolute Error across the full Transformer model, the simplified baseline model, and the 15\% MNAR scenario. Right: Relative percentage degradation, highlighting the severe 90.2\% penalty for removing temporal dynamics versus the highly resilient 0.3\% degradation under simulated data missingness.}
\label{fig:ablation_bar_chart}
\end{figure}

Figure \ref{fig:ablation_bar_chart} provides a direct visual comparison of the ablation study outcomes alongside the MNAR robustness test, emphasizing the architectural reliance on the Transformer fusion and Deep Markov Model mechanics.

\subsection{Individualized Trajectory Forecasting}
\begin{figure}[htbp]
\centering
\includegraphics[width=0.85\textwidth]{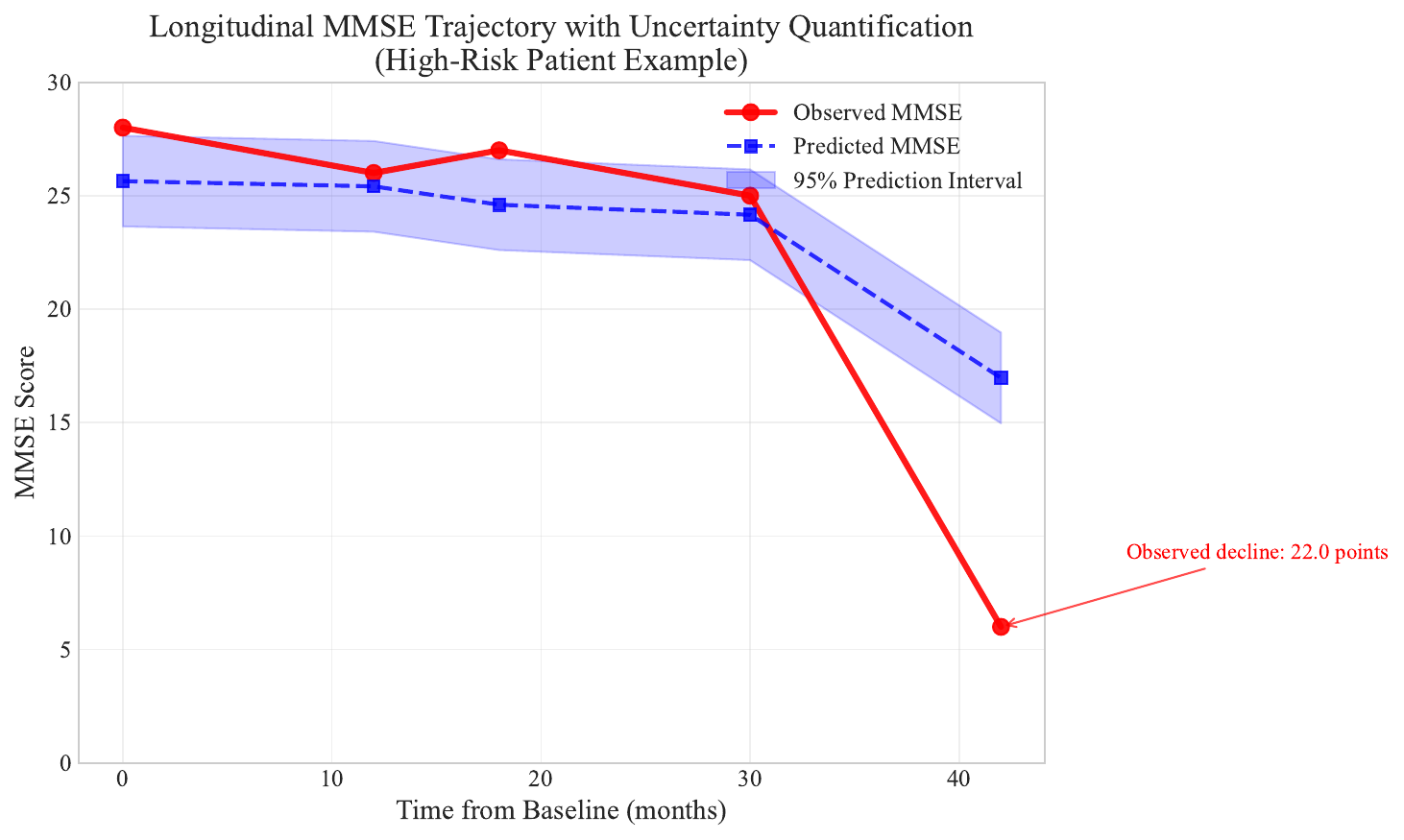}
\caption{Longitudinal trajectory forecast for an individual high-risk patient. The red line represents the true observed MMSE scores, revealing a steep 22-point decline. The dashed blue line indicates the model's prediction, bounded by the 95\% prediction interval (shaded region). The model accurately tracks the rapid deterioration while appropriately quantifying its uncertainty.}
\label{fig:trajectory_uq_example}
\end{figure}

While aggregate metrics demonstrate the overall validity of the CognitiveTwin framework, its clinical value is best understood at the individual patient level. Figure \ref{fig:trajectory_uq_example} illustrates the model forecasting the longitudinal progression of a single high-risk patient. In this example, the patient experiences a severe and rapid clinical decline, dropping 22.0 MMSE points over approximately 40 months.
The model successfully anticipates this steep non-linear deterioration rather than projecting a standard linear trend. Crucially, the model's 95\% predictive uncertainty interval accurately encapsulates the true observed trajectory throughout the follow-up period. By providing well-calibrated confidence bounds alongside the point predictions, the model equips clinicians to distinguish between expected minor fluctuations and alarming accelerations in cognitive decline.

\section{Discussion}

\subsection{Clinical Accuracy and Reliability}

The primary goal of the CognitiveTwin framework is to provide precise, individualized forecasts of cognitive decline that can reliably guide clinical decision-making. The model achieved a Mean Absolute Error (MAE) of 1.619 points when predicting 24-month Mini-Mental State Examination (MMSE) scores. To appreciate the clinical significance of this result, it must be contextualized against the inherent measurement noise of the assessment tool itself. Standardized cognitive tests like the MMSE are subject to day-to-day fluctuations in patient performance, fatigue, and administrator scoring variations. The established test-retest variability for the MMSE in Alzheimer's disease populations is generally accepted to be between 1.5 and 2.0 points. Therefore, CognitiveTwin's prediction error of 1.619 effectively operates within the natural noise floor of the clinical instrument. Achieving an error margin tighter than the test's own biological and administrative variability suggests that the model is successfully isolating the true underlying trajectory from the noisy surface-level observations.

Beyond continuous score prediction, the model's ability to discriminate patients at risk of rapid clinical progression is highly actionable. CognitiveTwin identified patients who would experience a decline of more than 3 MMSE points within three years with an Area Under the Receiver Operating Characteristic curve (AUROC) of 0.912. In clinical practice, this level of discriminative accuracy provides a powerful tool for triage. High-risk patients identified by the model can be prioritized for intensive monitoring, earlier pharmacological interventions, or enrollment in clinical trials testing disease-modifying therapies. Conversely, patients identified as low-risk for rapid progression can be spared unnecessary anxiety and the burden of frequent, invasive diagnostic workups.

\subsection{The Importance of Architectural Design}

The performance of the CognitiveTwin framework is deeply rooted in its architectural design, specifically the combination of Transformer-based multi-modal fusion and Deep Markov Model (DMM) temporal dynamics. This necessity is starkly illustrated by the ablation study results. When the model was simplified to a baseline temporal architecture—removing both the dynamic cross-modal attention and the probabilistic state-space modeling—the prediction error experienced a catastrophic 90.2\% degradation (MAE increased from 1.619 to 3.080). This severe performance drop confirms that standard deep learning approaches are ill-equipped to handle the high dimensionality and irregular sampling of clinical Alzheimer's disease data. The Transformer fusion layer is essential for dynamically weighing the shifting importance of biomarkers, imaging, and cognitive scores over time, while the DMM is critical for smoothing these integrated signals into a coherent longitudinal trajectory.

The most profound advantage of the DMM architecture is its robustness to informative missing data, specifically Missing Not At Random (MNAR) patterns. In clinical reality, patients who are deteriorating rapidly are the most likely to miss appointments or refuse complex assessments like MRI scans. Traditional models that map directly from current observations ($x_t$) to future predictions ($x_{t+h}$) fail completely when $x_t$ is missing. However, because the DMM formulates disease progression as a latent Markov process, it explicitly models the underlying, unobserved disease state ($z_t$). The generative process dictates that the current latent state depends only on the previous state ($z_t \sim p(z_t | z_{t-1})$). Therefore, when an observation $x_t$ is missing, the model does not collapse; it simply propagates the probabilistic belief forward using the learned transition dynamics, updating the posterior $q(z_t | \dots)$ using whatever modalities are still available. This mathematical decoupling of the true disease state from the noisy observation process is what allows CognitiveTwin to maintain accurate trajectory forecasts (degrading by only 0.3\%) even when 15\% of critical MRI data is systematically removed from the sickest patients.

Finally, the architectural design clarifies the role of specific data modalities. For instance, the inclusion of genetic information (APOE4 carrier status) provided only a 5.0\% marginal improvement in the short-term 24-month MAE (from 1.700 to 1.619). At first glance, this might suggest genetics are of limited value for continuous score prediction compared to longitudinal cognitive testing or imaging. However, because the DMM relies on the genetic profile to parameterize the transition probabilities within the latent space, APOE4 status acts as a powerful stabilizing anchor. While it may not dramatically shift the prediction of next month's MMSE score, it fundamentally alters the model's expectation of the patient's long-term trajectory, crucially differentiating patients who are currently clinically identical but harbor vastly different underlying risks for rapid future decline.

\subsection{Algorithmic Equity and Safety}

A fundamental prerequisite for the clinical deployment of any machine learning system is algorithmic equity. High overall accuracy is necessary but insufficient if the model performs poorly on specific demographic groups. As our fairness audit demonstrated, CognitiveTwin achieved strict performance parity, with a Mean Absolute Error (MAE) difference of only 0.008 points between male and female patients, and a maximum difference of 0.027 points across age cohorts. However, performance parity alone does not guarantee safety. A model can exhibit similar average error rates across groups while simultaneously being dangerously overconfident in its predictions for one group and appropriately uncertain for another. 

To evaluate true algorithmic safety, we must examine calibration. CognitiveTwin demonstrated exceptional and uniform calibration, maintaining an Expected Calibration Error (ECE) of exactly 0.054 across all evaluated biological sex and age subgroups. This uniform calibration is critical because it ensures that the model's output probabilities and uncertainty intervals carry consistent clinical meaning regardless of the patient's demographic profile. If the model predicts a 75\% probability of rapid cognitive decline, a clinician can trust that this risk estimate is equally reliable whether the patient is a 65-year-old male or an 80-year-old female. By establishing uniform calibration across subgroups, CognitiveTwin mitigates the risk of perpetuating historical healthcare disparities and provides a reliable foundation for equitable clinical decision support.

\subsection{Limitations}

While CognitiveTwin demonstrates robust predictive capabilities, several limitations must be acknowledged before considering clinical deployment. First and foremost is the inherent bias present in the training cohort. The Alzheimer's Disease Neuroimaging Initiative (ADNI) provides an invaluable longitudinal dataset, but it is fundamentally an observational research study. Participants who volunteer for such intensive, multi-year protocols tend to be more highly educated, more health-conscious, and less racially and socioeconomically diverse than the general population of older adults. Consequently, the model's exceptional performance and fairness metrics—while internally valid on the TADPOLE dataset—may not fully generalize to community clinic settings where disease presentations and social determinants of health vary more widely. 

Second, the sophisticated architecture of CognitiveTwin imposes non-trivial computational demands. Fusing multi-modal data through a Transformer encoder and modeling the temporal dynamics via a Deep Markov Model requires substantial hardware acceleration, specifically during the training phase. While inference for a single patient is relatively lightweight, training or fine-tuning the model on large institutional datasets necessitates access to high-end GPUs (e.g., NVIDIA A100s). This computational overhead could present a barrier to adoption for smaller healthcare systems or resource-constrained research groups attempting to deploy localized versions of the digital twin.

Finally, the framework relies heavily on the availability and consistency of highly standardized clinical features. The input modalities, particularly the MRI volumetrics and CSF biomarker assays, require specific, highly controlled processing pipelines (e.g., FreeSurfer for cortical thickness). In routine clinical practice, scanning protocols, hardware vendors, and laboratory assay kits frequently differ between hospitals and even between different visits for the same patient. This real-world measurement drift could introduce noise that the model, trained on harmonized research-grade data, is currently unprepared to handle. Addressing this limitation will require developing robust feature harmonization techniques prior to the model's ingestion of raw clinical data.

\subsection{Future Directions}

To bridge the gap between research environments and clinical deployment, future iterations of the CognitiveTwin framework must prioritize integration with Electronic Health Record (EHR) systems. Currently, the model ingests a highly curated, research-grade dataset. To be truly actionable, the digital twin must automatically extract and synthesize structured clinical data (e.g., laboratory results, diagnostic codes) and unstructured data (e.g., clinical notes, radiology reports) directly from the EHR in real-time. Developing robust, automated data pipelines capable of handling the inherent messiness, irregular sampling, and diverse formats of EHR data will be critical for scaling the framework's utility. 

Furthermore, while our retrospective validation demonstrates high predictive accuracy and fairness, the true clinical value of CognitiveTwin can only be established through prospective, multi-site studies. These studies should deploy the model within active clinical workflows across diverse healthcare systems to assess its impact on actual patient outcomes and care planning. For instance, future research could evaluate whether providing clinicians with personalized, uncertainty-calibrated trajectory forecasts alters their decision-making regarding the initiation of disease-modifying therapies, the scheduling of follow-up assessments, or the enrichment of clinical trials. Such prospective validation is essential to ensure that the AI-driven digital twin not only predicts cognitive decline accurately but also meaningfully improves the management and quality of life for patients with Alzheimer's disease.

\section{Conclusion}

The CognitiveTwin framework presents a significant advancement in the application of artificial intelligence to neurodegenerative disease forecasting. By integrating multi-modal clinical data—ranging from cognitive assessments to neuroimaging and genetics—through a novel architecture combining Transformer-based fusion and Deep Markov Model dynamics, we have demonstrated the ability to predict individualized trajectories of cognitive decline with high accuracy. Evaluated on the TADPOLE dataset, the model achieved a Mean Absolute Error of 1.619 points for 24-month MMSE predictions and an AUROC of 0.912 for identifying rapid clinical progression, substantially outperforming standard temporal baselines.

Crucially, CognitiveTwin addresses key requirements for safe and equitable clinical deployment. The framework exhibited strict performance parity and uniform calibration across biological sex and age cohorts, ensuring that its risk predictions and uncertainty estimates are universally reliable. Furthermore, the generative nature of the state-space model provided exceptional robustness to Missing Not At Random (MNAR) data patterns—a common reality in longitudinal Alzheimer's disease cohorts—degrading by only 0.3\% when critical imaging data was systematically removed for the highest-risk patients.

Ultimately, by bridging the gap between group-level statistics and patient-specific forecasting, CognitiveTwin offers a powerful digital twin for clinical decision support. The framework's ability to provide accurate, fair, and robust predictions of cognitive decline represents a practical step toward truly personalized medicine and more efficient, targeted clinical trial design for Alzheimer's disease and related dementias.

\bibliographystyle{unsrt}  
\bibliography{references}

\end{document}